\documentclass[journal]{IEEEtai}  
\usepackage{amsmath,amsfonts,amssymb}
\usepackage{graphicx}
\usepackage{hyperref}

\title{Small Object Detection in Industrial Recycling: A New Dataset and YOLO Performance Evaluation}

\author{\IEEEauthorblockN{Oussama Messai\IEEEauthorrefmark{1}*,
Abbass Zein-Eddine\IEEEauthorrefmark{1},
Abdelouahid Bentamou\IEEEauthorrefmark{1},
Mickaël Picq\IEEEauthorrefmark{2},
Nicolas Duquesne\IEEEauthorrefmark{3},
Stéphane Puydarrieux\IEEEauthorrefmark{3},
and Yann Gavet\IEEEauthorrefmark{1}}\\[6pt]
\IEEEauthorblockA{\IEEEauthorrefmark{1}Mines Saint-Etienne, CNRS, UMR 5307 LGF, F-42023 Saint-Etienne, France}
\IEEEauthorblockA{\IEEEauthorrefmark{2}Siléane Group, 12 Rue Louis Soulier, 42000 Saint-Etienne, France}
\IEEEauthorblockA{\IEEEauthorrefmark{3}Orano Group, Orano Recyclage La Hague -- 50444 La Hague Cedex, France}}

\begin{document} 
\maketitle

\begin{abstract}
In this paper, we address the problem of detecting small, dense, and overlapping objects, a major challenge in computer vision. Our focus is on reviewing proposed methods based on deep learning supervised approaches. We provide a detailed comparison of these systems on a new dataset of more than 10k images and 120k instances, highlighting their performance, accuracy, and computational efficiency in the industrial recycling process use case. Through this comparative analysis, we identify the most reliable systems currently available and the specific challenges they are designed to tackle. Furthermore, we explore the benefits of data augmentation and synthetic images. Based on our analysis, we also propose potential future directions and innovative solutions that could enhance the effectiveness of small, dense and overlapped object detection systems. The scope of our investigations encompasses object detection, length measurement, and anomaly detection within the context of the recycling process. The anomaly detection strategy is robust against variations in image resolution and zoom levels, ensuring reliable performance in industrial applications.
\textit{The repository of the proposed dataset, methods and evaluation codes can be found at:  \href{https://github.com/o-messai/SDOOD}{https://github.com/o-messai/SDOOD}}
\end{abstract}


{\noindent \footnotesize\textbf{*} Address all correspondence to Yann Gavet, 
École des Mines de Saint-Étienne, 158 Cours Fauriel, 42023 Saint-Étienne, France. 
Email: \href{mailto:gavet@emse.fr}{gavet@emse.fr}, 
Tel: +33 (0)4 77 42 01 23}

\section{Introduction}
\label{sec:intro}
Object detection algorithms are a critical technology in many industrial applications, where they enable automation, quality control, and improved decision-making processes \cite{zhao2019object}. While significant advancements have been made in object detection algorithms through the application of deep learning techniques, the detection of small and overlapping objects remains a particularly challenging task \cite{rabbi2020small}. Industries such as manufacturing, robotics, logistics, and quality assurance, where precise identification and tracking of items are crucial for efficiency and safety, stand to benefit greatly from improved solutions to this problem. In these settings, the ability to accurately detect small objects that may be partially obscured or overlapping is crucial for tasks such as assembly line monitoring, inventory management, and defect detection \cite{messaiQCAV25}.

\begin{figure}[ht]
    \centering
    \includegraphics[width=\linewidth]{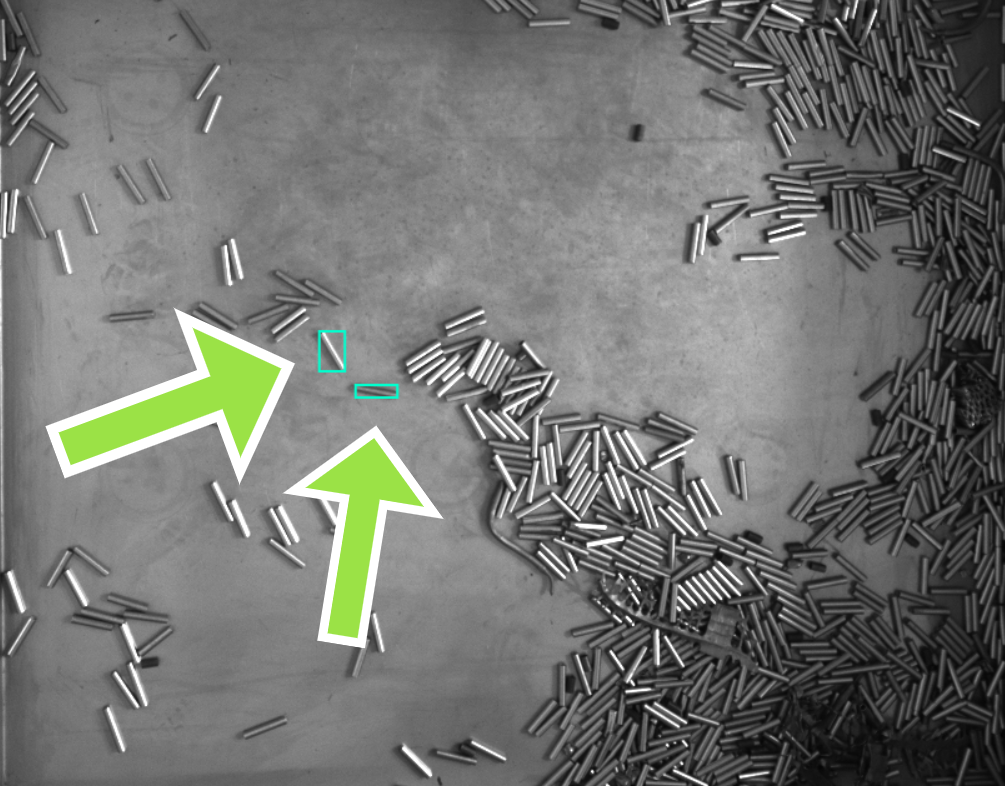}
    \caption{Example of small, dense and overlapped objects in an industrial recycling process. Arrows indicate bounding box detection, which is expected from an object detection system.}
    \label{fig:intro}
\end{figure}

Industrial environments often include components, parts, or products that need to be identified and tracked individually, even when they are closely packed or partially occluded \cite{pathak2018application}. For example, in automated manufacturing, small components may need to be detected on a conveyor belt where they might overlap or be closely clustered. Failure to detect these objects accurately can lead to operational inefficiencies, increased error rates, and potential safety hazards. In quality control processes, overlapping objects such as small electronic components on a circuit board need precise identification to ensure that assembly meets stringent standards without defects or misplacements. 

Detecting small, dense, and overlapping objects poses significant challenges in computer vision as  shown in figure \ref{fig:intro} due to the complex nature of these scenes, where objects exhibit high variability in scale, position, and occlusion \cite{zou2023object}. Industrial applications often involve scenarios with objects that are not only small but also densely clustered and overlapping, making accurate detection both critical and difficult. In such environments, detection systems are prone to higher rates of false positives/negatives and missed detections. When objects overlap, parts of one or more objects may be obscured, making it difficult for the detection algorithm to accurately identify and delineate the individual objects. These challenges lies in the limitations of both traditional and machine-learning based object detection algorithms, which often struggle with distinguishing features and maintaining accuracy under such conditions. In the following, we describe the most challenges of overlapping objects scenario:  

\begin{itemize}
    \item Bounding Box Accuracy: Overlapping objects can lead to inaccuracies in the placement and size of bounding boxes. The algorithm might struggle to correctly separate the objects, leading to merged or misaligned bounding boxes.
    \item Class Confusion: Overlapping objects may confuse the classification process. The algorithm might misclassify the objects due to mixed features from the overlapping regions, especially if the objects belong to different classes.
    \item Feature Extraction Challenges: Overlapping objects complicate feature extraction as the features of one object can interfere with those of another. This interference can reduce the overall effectiveness of feature detection and object recognition.
\end{itemize}

In addition, many industrial processes require the detection of small objects. In the following, we describe the most common challenges of small objects scenario: 

\begin{itemize}
    \item Reduced Feature Representation: Small objects often occupy a very small number of pixels, leading to a limited representation of their features. This makes it difficult for the detection algorithm to extract meaningful features, which can result in missed detections or incorrect classifications.
    \item Scale Variability: Detecting small objects requires handling significant scale variations effectively. Algorithms may struggle to maintain accuracy across different scales, especially when the objects are much smaller than the average objects seen during training.
    \item Anchor Box Limitations: In anchor-based detection methods, predefined anchor boxes may not be optimally suited for detecting small objects, leading to poor overlaps between the anchor boxes and the actual objects, thereby affecting detection performance.
    \item Sensitivity to Resolution Changes: Small objects are highly sensitive to changes in image resolution. Any reduction in resolution or downsampling can cause small objects to disappear or become indistinguishable, severely affecting detection accuracy.
\end{itemize}

Regarding the noted challenges, there is a pressing requirement for better object identification systems that can effectively recognize and distinguish small, overlapping objects in real time. Addressing this demand not only improves the precision and reliability of industrial processes but also increases overall productivity, minimizes waste, and ensures greater levels of product quality and safety. In this study, we focus on a deep learning-based approach to the recycling process. We investigate future perspectives and potential solutions, providing insights based on our comprehensive research. This paper is intended to guide future research and development efforts toward more robust and reliable dense overlapping object identification systems. This work includes the following main contributions:

\begin{enumerate}
    \item A new dataset with over 10,000 images and 120,000 instances labeled manually. A new case study that demonstrates the practical uses of standard object detection models in industrial recycling processes. 
    \item A comprehensive evaluation and comparison of the most common deep learning-based object identification systems, with an emphasis on their performance in dense and overlapped object recognition.
    \item An in-depth pre-processing study aimed at improving object detection performance, as well as an assessment of the precision and reliability of length measurements using simply bounding box coordinates.
    \item We evaluate the reliability of synthetic datasets for training supervised object detection models for industrial recycling process application. Additionally, we conduct tests of the model through simulations of its deployment in real-world scenarios.
    \item We propose an anomaly detection strategy which adapts effectively to fluctuations in image resolution and zoom factors.
\end{enumerate}

This study is located within the industrial context of Orano Group, a globally renowned leader in nuclear fuel production and recycling. Its facility of La Hague is tasked with processing irradiated nuclear fuel from various power plants, extracting reusable materials such as uranium and plutonium, while ensuring the safe management and storage of radioactive waste. The recycling process at Orano group industries is a cornerstone of the facility’s operations, and its optimization is critical to the future of the nuclear energy sector. One of the main objectives of this project is the detection, qualification, and measurement of the length of the metallic hulls during the initial stages of the recycling process (see Fig. \ref{fig:intro}).

\section{New Dataset}
\label{sec:Dataset}

\begin{figure}[ht]
    \centering
     \includegraphics[width=0.6\linewidth]{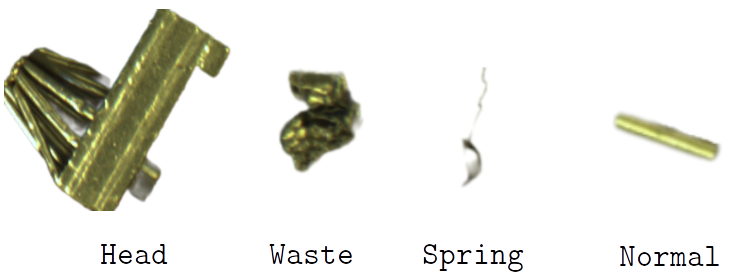}
     \caption{Visual examples of the main object classes for detection in the recycling system.}
     \label{fig:seven_objects}
\end{figure}

Defining whether an object is small or not can be done using various metrics, often depending on the context and image resolution. In some cases, subjective or perceptual criteria might be applied, where objects are considered small based on human perception and practical considerations within the specific application context. However, the basic metric is percentage of bounding box area of the object related to the image area. Percentage of Image Area: One straightforward method is to define a threshold based on the percentage of the total image area that the object occupies \cite{tamiminia2020google}. The percentage in pixels can be obtained by the following simple formula: \\

\begin{equation}
    Percentage_{\text{Area}}=(\frac{\text{Bounding Box Area}}{\text{Image Area}} )\times100
\end{equation}
For example, an object might be considered small if it occupies less than a certain percentage of the total image area (e.g., 1\%, 5\%).

The choice between actual and synthetic datasets for training object detection models is influenced by a number of parameters, including the unique application needs, data availability, and desired level of training control. The following sections discuss the actual and synthetic datasets used in our study.

\subsection{Real images}

\begin{figure}[ht]
    \centering
    \includegraphics[width=0.39\linewidth]{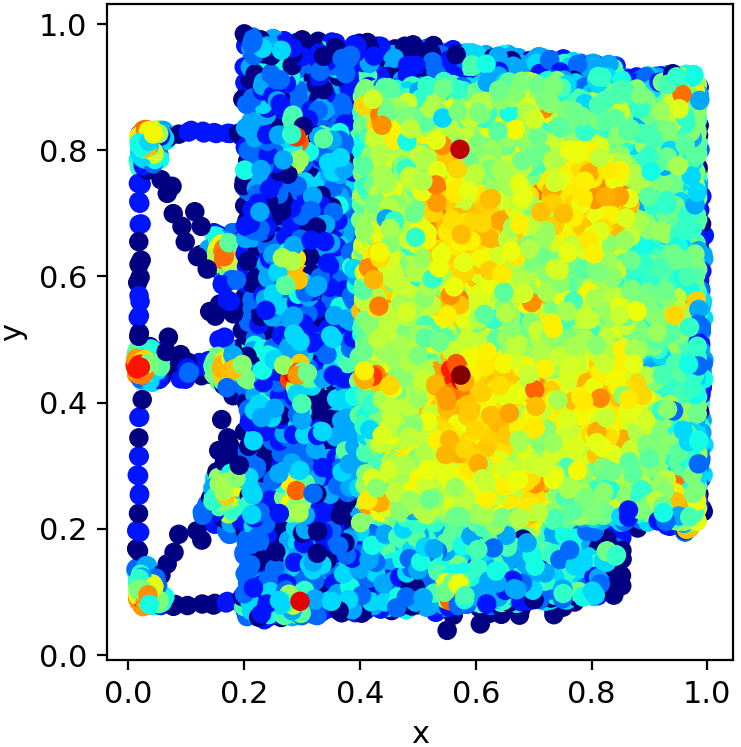}
    \includegraphics[width=0.39\linewidth]{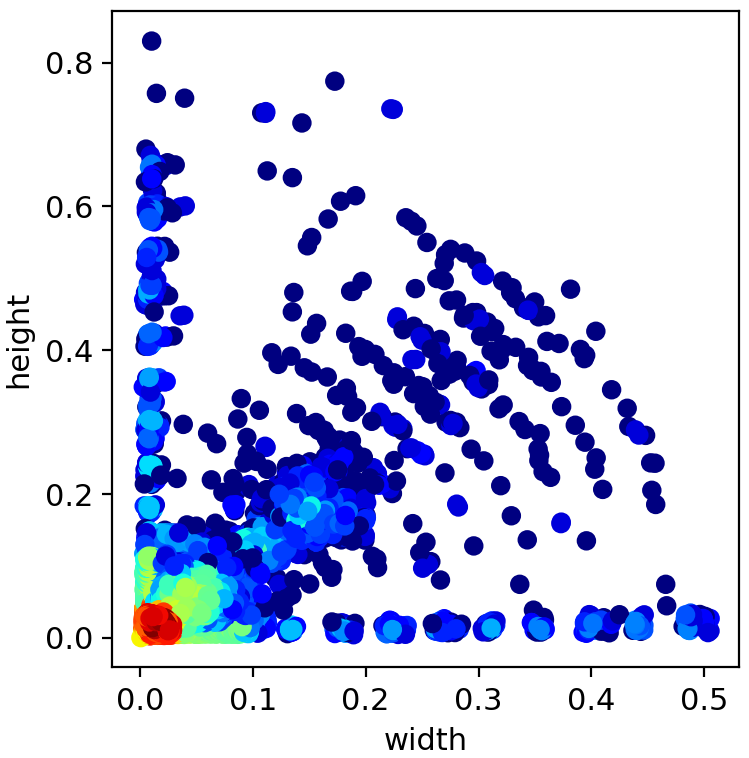}
    \caption{Heatmap showing the distribution of bounding box annotations across all classes, where (x, y) represent the center coordinates of the bounding box, and h, w correspond to its height and width.}
    \label{fig:heatmap_auth}
\end{figure}

It is well established that authentic real-world images yield superior performance, as they closely resemble actual use cases. However, synthetic images have also proven to be reliable and capable of delivering excellent results. This study introduces a dataset containing over 10,000 authentic images and approximately 120,000 densely overlapped, small objects. The dataset includes 7 industrial object classes relevant to recycling applications, including different metallic hull variations (normal, deformed, pinched, smashed), and other objects such as springs, waste, and heads. Figure \ref{fig:seven_objects} provides visual examples of these classes. 
The estimated sizes of the object classes, along with their detection objective difficulty scores (ranging from 1, easy, to 6, hard), are provided in Table \ref{tab:my_classes}, note that the object size is correlated the detection difficulty level from a human perspective. Most images have a resolution of 2448 x 2048 pixels, with an average of 12 objects per image, a minimum of 1 object, and a maximum of 986 objects in a single image. All annotations were manually conducted by humans, with the overall distribution of annotations presented in Fig. \ref{fig:heatmap_auth}. As illustrated in the figure, most object heights and widths are concentrated near zero, while their positions are distributed throughout the scene.

\begin{table}[ht!]
\centering
\resizebox{\columnwidth}{!}{%
\begin{tabular}{|c||c|c|c|}
\hline
 ID  &  Class & Obj. Difficulty level & Size in \% \\
\hline                     
         0&  Normal & 3 & $\approx 0.01 \% $\\
         1&  Deformed  & 3 & $\approx 0.02 \% $\\
         2&  Pinched  & 5 & 	$\approx 0.01 \% $\\
         3&  Smashed  & 4 & $\approx 0.01 \% $\\
         4&  Spring   & 6 & $\approx 0.01 \% $\\
         5&  Waste   & 2 & $\approx 0.5 \% $\\
         6&  Head  & 1 & $\approx 2.5 \% $\\    
\hline
\end{tabular}}
\caption{Object class size is expressed as a percentage, along with the corresponding level of difficulty in object detection. Detection objective difficulty ratings, ranging from 1 (easy) to 6 (hard).}
    \label{tab:my_classes}
\end{table}


\subsection{Synthetic images}

\begin{figure}[ht]
    \centering

     \includegraphics[width=0.28\linewidth]{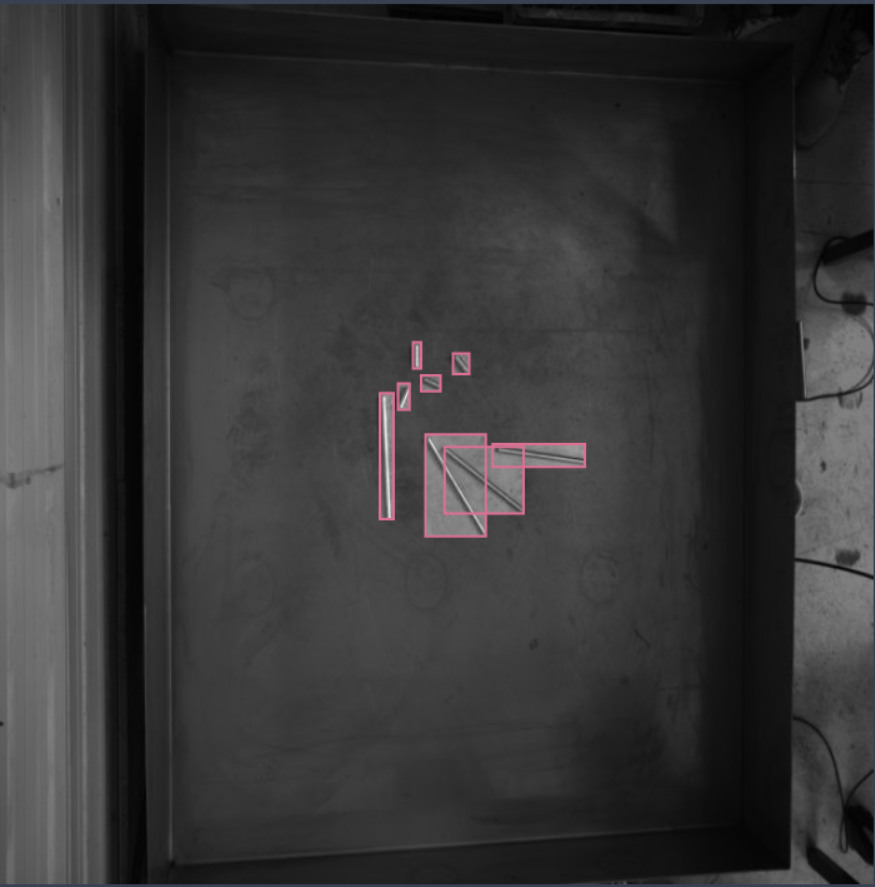}
     \includegraphics[width=0.28\linewidth]{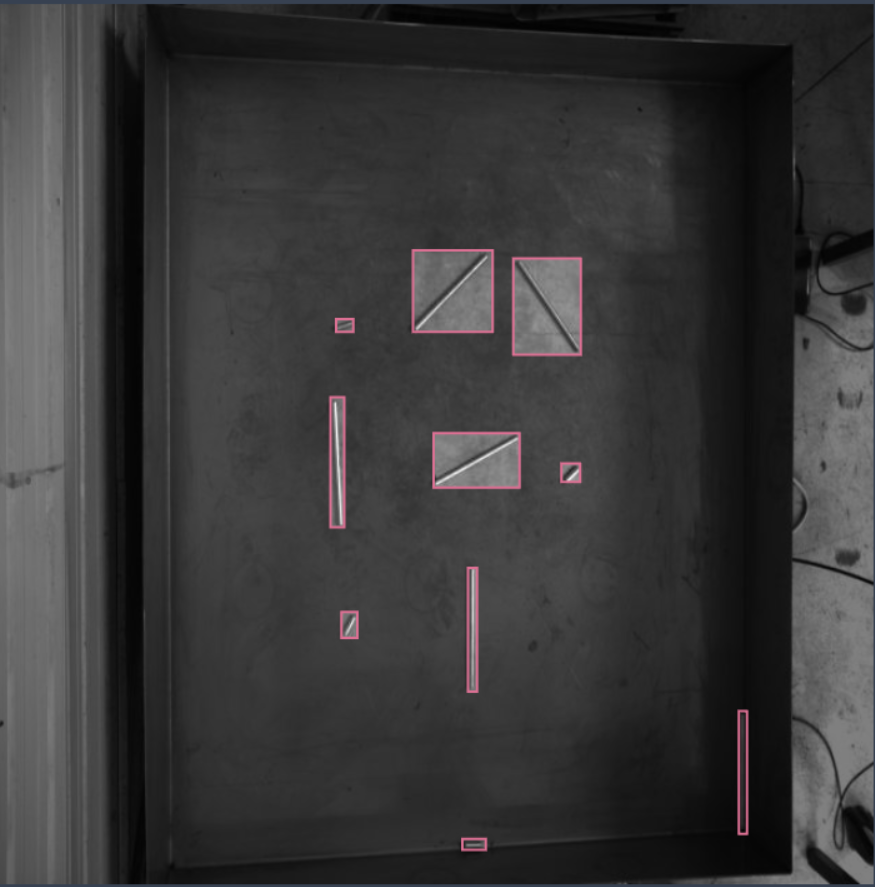}
     \includegraphics[width=0.28\linewidth]{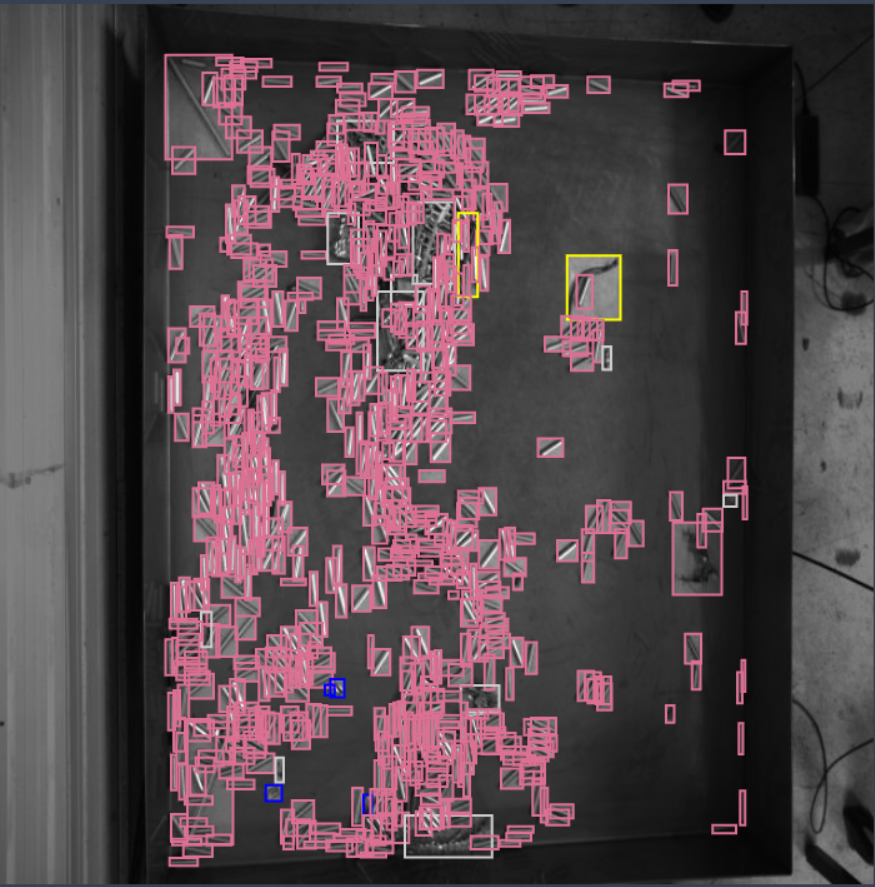} \\
     
     \includegraphics[width=0.28\linewidth]{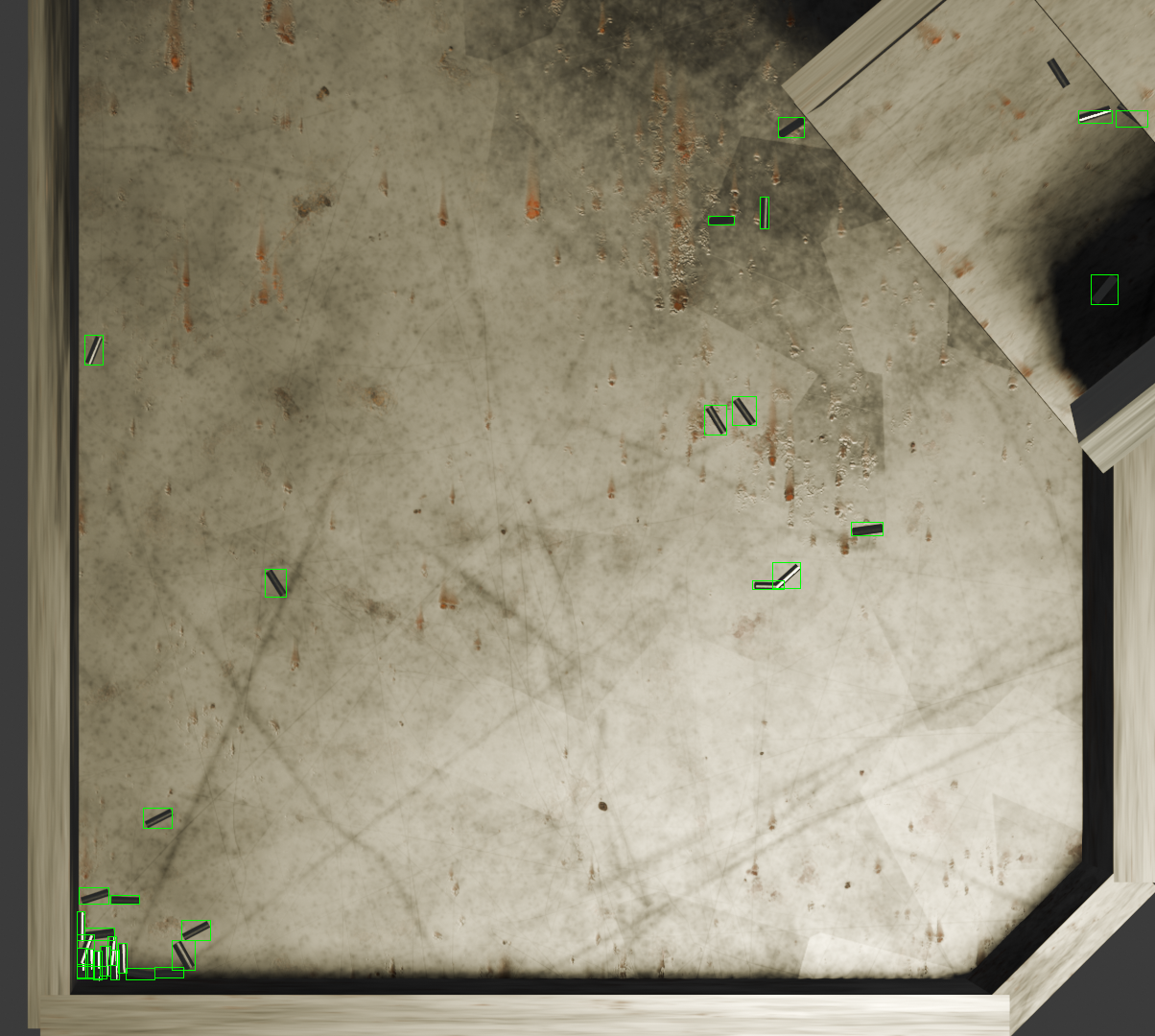}
     \includegraphics[width=0.28\linewidth]{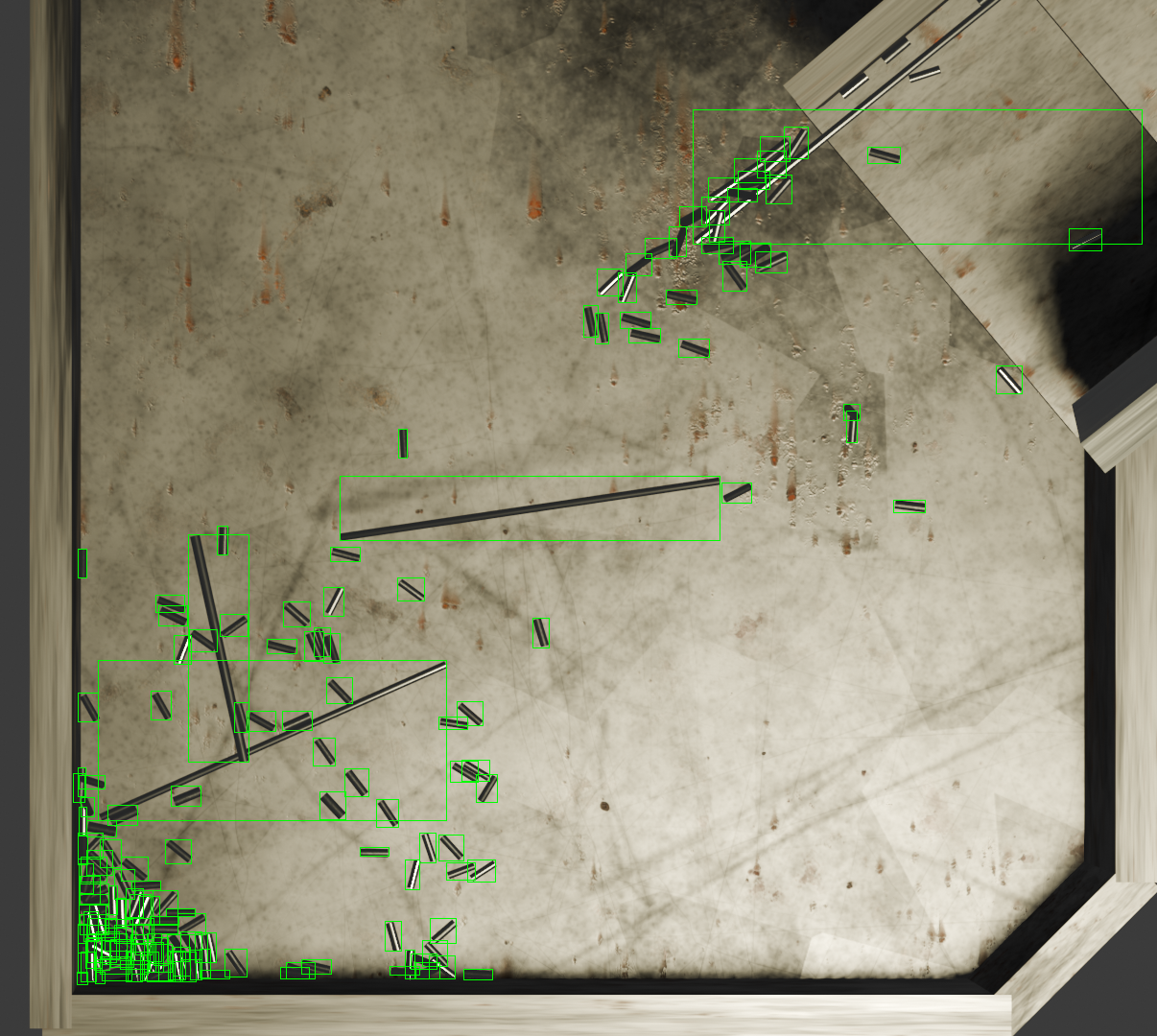}
     \includegraphics[width=0.28\linewidth]{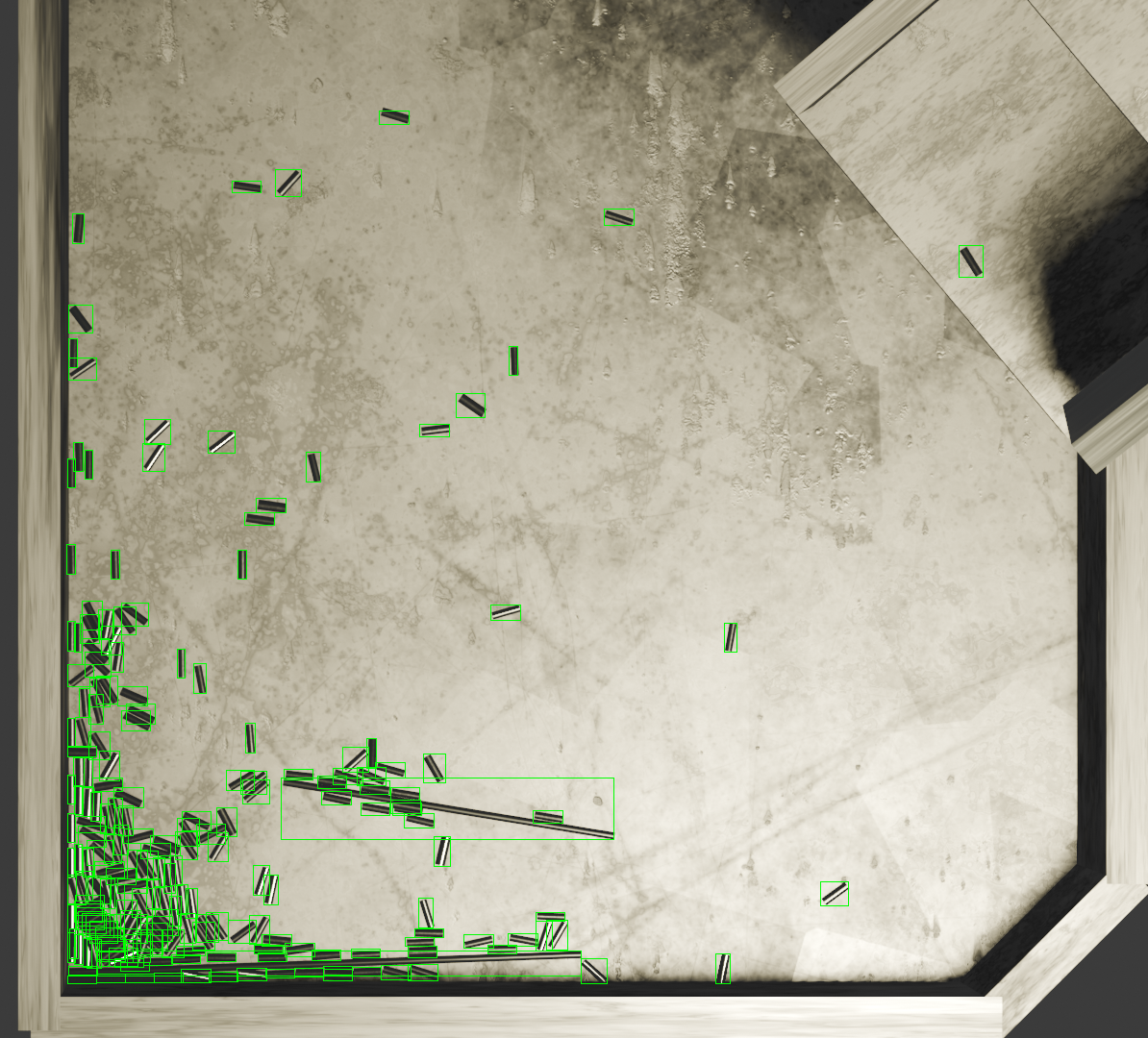} 
    \caption{Examples of labeled real and synthetic images. Top: real images, bottom: synthetic images.}
    \label{fig:real_fake_examples}
\end{figure}

The real images inherently reflect real-world conditions, providing a more realistic training environment for detection models, the collection of actual data particularly in scenarios involving small, dense, and overlapping objects can be both costly and time-consuming. Furthermore, the annotation process poses significant challenges due to the intricacies involved in accurately labeling these objects. The images were managed and annotated using the Roboflow online software \cite{dwyer2022roboflow}.

Meanwhile, synthetic datasets can be generated on a large scale, allowing for complete control over object location, density, and overlap as well as automatic labeling. However, in cases where object features are complex, the creation of synthetic images can become more challenging and costly. 
We generated a synthetic dataset comprising approximately 50,000 images and 6.5 million instances using the well-known open-source software Blender \cite{rohe2022generation}. This dataset focuses on a single class (i.e., the normal class), as modeling and generating other object structures is time-consuming. The primary objective of the synthetic data is to investigate its impact on training with real data and to assess its reliability in enhancing performance, as well as the effects of adding one class compared to others. Examples of real and fake images are shown in Fig. \ref{fig:real_fake_examples}.

\section{RELATED WORK}
\label{sec:Proposed_method}
In this study, we benchmark real-time and near real-time object detection methods based on supervised deep learning techniques. Additionally, we briefly discuss non-deep learning-based approaches, covering traditional image processing methods, deep learning-based techniques, and hybrid approaches that combine elements from both categories.

\subsection{Image processing based approaches}
\label{subsection:image_processing}

Object detection methods based on image processing have a historical precedence that predates the emergence of deep learning techniques \cite{gupta2019improved}. These traditional methods rely on hand-crafted features and mathematical models to analyze and interpret visual data, forming the foundation of early advancements in computer vision. However, such approaches often face limitations when dealing with complex scenarios, such as detecting small or overlapping objects, where feature extraction becomes more challenging. Authors in \cite{ahmad2011geometric} developed a method to fit polygonal shapes (rectangles or prisms) around overlapped ammonium oxalate crystals in pure water. This method is composed of two main steps. The first step is to detect salient corners, where each corner is represented by its x, y coordinates and the two direction vectors of the two lines intersecting at that corner. To find those salient corners, some pre-processing is applied on the studied image. First, the background is removed, before applying Rosten and Drummond \cite{rosten2005fusing} corners detector to generate a set of candidate corners. Next, the contours are extracted using multi-scale Canny edge detector \cite{canny1986computational} here the multi-scaling is achieved by changing the standard deviation of the Gaussian kernel applied in the Canny edge detection process. The authors applied some morphological filters to remove noise and the internal edges that are obtained, other operations such as closing and convex are used. From the candidate corners, the salient corners are selected. The salient corners are then clustered in a way to get at the end the three corners that represent one object (rectangle or prism). 
In \cite{de2016multiscale} another method is proposed to recognize overlapped elliptical particles in 2D images. the developed method is based on three main steps,  first is a binarization of the input image for contour detection and segmentation where a threshold is applied. Second, a median filter is applied to remove noise before grouping the segments. Third is the decomposition of the cluster. The contour segments are detected using the well known approach 'connecting point' \cite{honkanen2005recognition}. The method was tested on 30 synthetic images of overlapped elliptical objects where it was able to detect up to 91.3 \% of the objects. In addition it was also tested on real images both noisy and non-noisy and the results illustrate high accuracy and robustness. However another similar approach has been proposed in \cite{de2018efficiency}, where several improvements are highlighted, first a new ratio is introduced to discriminate a single ellipse from clusters of two ellipses with single concavity point. Second, in each cluster, all the possible combinations of the segments are identified. The comparison was done on both synthetic data and real data, where the method shown remarkable accuracy improvement.

\subsection{Deep learning based approaches}
\label{subsection:deep_learning}
Deep learning approaches have shown promise in addressing the challenges of overlapped object detection. These approaches typically involve training a Convolutional Neural Network (CNN) on a large dataset of images and corresponding object annotations rather than relying on hand-designed features, the model is trained to predict the locations and categories of objects from end-to-end. However, these types of models also struggle to detect small and overlapping objects, where the challenge lies in accurately recognizing object shapes and identities. Overall, object detection models can be classified into two main categories: Single-Stage Detectors, these models perform object detection in a single step by directly predicting both bounding boxes and class probabilities from the input images. They are generally faster and more suited for real-time applications. The second category is called Two-Stage Detectors. In this approach, the detection process is divided into two stages. The first stage generates region proposals, identifying potential object locations, while the second stage classifies these regions and refines the bounding boxes. Although typically slower than single-stage detectors, two-stage models often achieve higher accuracy. For example, Fast RCNN (2015) \cite{girshick2015fast} use a pre-trained convolutional neural network to extract feature maps from the entire image at once. It then uses Region of Interest (RoI) pooling to extract information unique to each object proposition. The RoI technique aids in transforming features associated with numerous object proposals into fixed-size feature maps, enabling for efficient processing of object proposals at varied scales. Ren et al. \cite{ren2016faster} (2016) presented Faster R-CNN, which improves on Fast R-CNN by eliminating the necessity for an external region proposal technique like Selective Search. Instead, it adds a Region Proposal Network (RPN) into the design, rendering the detection pipeline completely convolutional and substantially faster. The RPN is a tiny network that shares convolutional features with the larger detection network. It scans the image with sliding windows and creates region recommendations based on predicted objectness scores and bounding box coordinates. Following the generation of region proposals, the RoI pooling layer is applied, and the object proposals are classed and refined (in terms of bounding box position) in the second stage.

\subsubsection{From YOLO-v1 2016 to YOLO-11 2024}
\label{sec:from_v1_to_v11}
The YOLO (You Only Look Once) family of models has been continuously improved to handle such scenarios better, there are specific approaches and modifications within and beyond YOLO that might be effective for small and overlapped object detection. The definition of a small object in general use cases: YOLO is a widely-used real-time object detection method known for its speed and accuracy. Over time, YOLO has undergone several iterations, each building on the performance of its predecessor. In the following, an overview of the significant versions and updates:

YOLOv1 (2016) \cite{redmon2016you}: The initial version introduced the concept of framing object detection as a regression problem, predicting bounding boxes and class probabilities directly from full images, enabling fast detection. Introduced the single-stage detection approach, where the model looks at the image only once to predict bounding boxes and class probabilities. The architecture was based on a single Fully Convoluation Network (FCN) that divided the image into a grid and predicted bounding boxes and probabilities for each grid cell.

YOLOv2 (2017) (also known as YOLO9000) \cite{redmon2017yolo9000}: Improved accuracy and speed over YOLOv1. Introduced anchor boxes to better predict bounding boxes of varying shapes and sizes. Incorporated batch normalization to improve training stability and convergence. Utilized high-resolution classifier and multi-scale training for better detection of small objects.

YOLOv3 (2018) \cite{redmon2018yolov3}: This version brought significant improvements in accuracy and detection of smaller objects by incorporating a multi-scale detection approach. YOLOv3 introduced Darknet-53 as its backbone network, which allowed better feature extraction. It also utilized three different scales for detecting objects, improving the performance on small, medium, and large objects. Additionally, YOLOv3 employed logistic regression for class prediction and used independent logistic classifiers to address the multi-label classification problem.

YOLOv4 (2020) \cite{bochkovskiy2020yolov4}: Significant improvements in both speed and accuracy. Implemented various techniques such as Cross-Stage Partial connections (CSP), Mish activation function, and a modified Path Aggregation Network (PAN).
Improved data augmentation methods like Mosaic augmentation and Self-Adversarial Training (SAT). Emphasized on balancing between model size, speed, and accuracy for real-world applications.

YOLOv5 (2020) \cite{jocher2022ultralytics}: Developed by Ultralytics, YOLOv5 prioritizes usability, speed, and versatility in object detection tasks. This version features a PyTorch-based implementation, which enhances accessibility and facilitates modifications for users. It offers a range of pre-trained models, accompanied by comprehensive documentation for various application scenarios. 

YOLOv6, YOLOv7 (2022)  \cite{li2022yolov6,Wang_2023_CVPR}: Continued improvements in model efficiency and accuracy. Further optimizations for deployment in various environments, including mobile and edge devices. YOLOv7 introduced new techniques such as E-ELAN (Efficient and Lightweight Attention Network) and additional training techniques. The E-ELAN enhances feature representation and processing efficiency while maintaining low computational costs. By emphasizing key features in the input data, it achieved state-of-the-art performance regarding speed and accuracy at the time of its release.

YOLOv8 (2023)  \cite{terven2023comprehensive}: Featuring advanced techniques for better performance. Emphasized modular design for easier customization and integration. Continued improvements in speed and accuracy, making it suitable for a wide range of applications from autonomous driving to security surveillance. The main key feature for this version is an innovative anchor-free split head, which improves both accuracy and efficiency compared to traditional anchor-based methods.

YOLOv9 (2024)  \cite{wang2024yolov9}: Building upon its predecessors, YOLOv9 enhances the architecture to achieve greater accuracy and speed in real-time object detection. Key enhancements consist of advanced feature extraction techniques that enhance the ability to detect small objects, as well as optimizations tailored for various hardware platforms. These advancements facilitate quicker inference times while maintaining high precision, making YOLOv9 particularly suitable for applications requiring rapid decision-making in dynamic environments.

YOLOv10 (2024) \cite{wang2024yolov10}: YOLOv10 represents the next evolution in the YOLO series, focusing on advanced network architecture and innovative training methodologies. This version introduces enhancements that allow for better handling of complex scenes and occlusions, thereby improving detection performance in challenging scenarios. Additionally, YOLOv10 emphasizes the integration of multi-modal data, enabling the model to leverage diverse input types for more robust detection capabilities. These developments position YOLOv10 as a state-of-the-art tool for real-time object detection across various applications.

YOLO11 (2024) \cite{yolo11_ultralytics}: a recent version is built upon YOLOv8 backbone and neck architectures. The Ultralytics team claims that this new model employs an anchor-free split head, enhancing both accuracy and efficiency compared to traditional anchor-based methods.

\subsubsection{Focusing on small, dense and overlapped objects}
\label{sec:focusing_on_small}
The demand for detecting small, dense, and overlapping objects in industrial applications has grown, particularly in quality control and automated inspection systems. Despite rapid advances in object detection models, a performance gap persists for recognizing dense and overlapped targets. Although various methods have been presented, the majority of the proposed models are variations or adjusted versions of the basic recommended detection systems. 

For example, Lou et al. \cite{lou2023dc} developed the DC-YOLOv8 method, which is specifically designed for detecting small objects utilizing camera sensors in industrial settings. The suggested architecture improves feature extraction to maintain context information, which is critical for detecting small and dense objects that frequently overlap with one another. The performance of the DC-YOLOv8 model demonstrates its promise in automated industrial inspection. Xiao et al. \cite{xiao2024efficient} proposed a new small-object detection framework to improve the accuracy of steel scrap quality inspection in congested situations. Their model improves identification precision by including a small-object detection layer and improving non-maximum suppression (NMS) strategies for dealing with overlapping items in cluttered situations. This technique indicates potential for industrial applications, where item size, density, and occlusion offer significant hurdles to existing detection algorithms. Noh et al. \cite{noh2024enhancing} improved the detection of small and densely packed objects using an adaptive non-maximum suppression (NMS) method designed for industrial applications. The Relation-NMS technique increases object detection by boosting context-awareness, allowing the system to efficiently handle high-density overlapping items commonly found in industrial processes such as quality control and inspection. Deng et al. \cite{deng2020global} suggested a global-local self-adaptive network that detects objects in drone-view photos. Their model focuses on instances in which small objects are closely packed, resulting in frequent overlaps. By adding a multi-scale approach and spatial context analysis, the network displays increased performance in recognizing densely grouped objects in real-world industrial applications like inventory management and automated monitoring. Wang et al. \cite{wang2024delving} discuss the problems of recognizing small, dense, and overlapped forbidden items during security inspections. They increase the efficiency in congested environments by improving data augmentation techniques. Djenouri et al. proposed an end-to-end CNN model similar to the YOLO model \cite{djenouri2020fast}, with a focus on industrial manufacturing applications and speed optimization. Authors in \cite{sun2021deep} present a solution for counting overlapping rice seeds on a conveyor belt based on Faster RCNN model prefixed by some classical image processing operations. The idea is to apply a pre-processing step for the image passed to the model using image processing techniques as a first step in addition to a modification in the Faster RCNN model to be adapted for the desired task.  In the pre-processing step, the rice image is binarized, and the outline of rice seeds are extracted. Then the centers of the rice seeds are obtained using fast radial symmetry transformation, followed by the contour grouping using the fusion comprehensive criterion of Euclidean distance and divergence function. Finally, the contour is reconstructed based on ellipse fitting method to pre-label the rice seed contour. 

Furthermore, in \cite{wang2021small}, an object detection system is suggested based on the concept of YOLOv3 \cite{redmon2018yolov3}. The authors suggest an improvement of speed and accuracy for small-object detection, they propose a deep learning small-object detection method. The main contribution of the proposed system is improving the image resolution before extracting object features. This pre-processing improves the resolution of the YOLO model, allowing it to detect content features of small objects more easily. Although that the system mainly focuses on small objects, it shows promising results in overlapping and fairly dense cases. Another YOLO-based model is proposed in the paper \cite{guo2022overlapped}, where the authors propose a detection system using YOLOv5 for the overlapped pedestrian use case. The authors of \cite{stewart2016end} proposed another method for detecting crowded pedestrian. The proposed detection approach works by first encoding the input image into high level descriptors using a CNN and then decoding that representation into a set of bounding boxes. This model is NMS-free and employs a Long Short Term Memory (LSTM) \cite{10.1162/neco.1997.9.8.1735} to predict bounding boxes sequentially. Using its ability to memorize, the LSTM will avoid recreating the same bounding box. The model is evaluated by two data sets the first one is Brainwash dataset and the second is TUD-Crossing dataset \cite{andriluka2008people}, and the results are compared to Fster-RCNN and YOLO models. Another work focusing on crowd pedestrian detection that proposes an NMS variant is presented in \cite{huang2020nms}. The method called Representative Region NMS (R$^{2}$NMS), leverages the visible parts of the pedestrians in NMS on highly overlapped full bodies. The main difference between the normal NMS and the (R$^{2}$NMS), is in the calculation of IoU. The latter calculates the IoU between the visible regions of the two bounding boxes instead of calculating it between the full-body boxes.

\section{EXPERIMENTAL RESULTS}
\label{sec:experimental}
\begin{table*}[ht!]
\centering
\resizebox{\columnwidth}{!}{%
\begin{tabular}{|c|c|c||c|c|c|c|c|}
\hline
    Model & Img size  &$mAP0.5-0.95_{COCO-17}$ & mAP@0.5 & mAP0.5-0.95 & Recall& Precision & Parameters \\
\hline                     
         YOLOv5x &$640^{2}$  &50.7 &37.7 &27.0 &35.0  &65.8 & 86.7M\\       
         YOLOv6l &$640^{2}$   &52.8  &49.8 &39.2 &55.6 &53.7 & 59.6M\\  
         YOLOv7x  &$640^{2}$   &53.1  &53.1  &38.0 &50.9 &\textbf{87.4}  & 71.3M\\ 
         YOLOv7-e6e  &$640^{2}$   &NA  &52.5   &37.5 & 50.4  &82.6  & 151.7M\\      
         
         YOLOv8m  &$640^{2}$  &50.2  &40.6 &26.6 &34.0 & 57.9 & 25.9M\\           
         YOLOv8l  &$640^{2}$  &52.9  &59.9 &45.0 &55.9 &80.6  & 43.7M\\ 
         YOLOv8x  &$640^{2}$  &53.9  &60.8 &45.6 &56.8 &80.7 &68.2M \\         
         YOLO-NAS-l  &$640^{2}$  &52.2  & 50.9&NA &\textbf{62.7}  &14.3 & NA\\
         YOLOv9-e  &$640^{2}$  &55.6 &57.9   &43.5 &51.9    &81.7   & 58.1M\\ 
         YOLOv10-x  &$640^{2}$  &54.4 & 48.3   &36.3 &48.5    &73.1   & 29.5M\\ 
         YOLO11-x  &$640^{2}$  &54.7  &\textbf{62.8}   &\textbf{46.7}  & 57.9   &81.9    &56.9M   \\ 
                         
         Faster R-CNN-R101  &$640^{2}$  &44.0 &50.1 &39.4 &61.2 &74.5 & 60M \\ 
         RT-DETR-l  &$640^{2}$  &53.0  &40.1 &27.9  &44.3 &69.1 & 32M \\
         RT-DETR-x  &$640^{2}$  &54.8  &48.9 &33.5 &50.1 &62.6  & 67M\\             
         \hline

         YOLOv5x6  &$1280^{2}$  &55.0  &55.0 &42.2   &45.9 &87.1 & 140.7M\\
         
         YOLOv6-l6  &$1280^{2}$  &57.2  &53.0 & 42.6 &50.9 &81.6& 140.4M \\       
         RT-DETR-l  &$1280^{2}$  &NA  &46.0 &33.5  &46.2 &69.2  & 32M \\ 
         RT-DETR-x  &$1280^{2}$  &NA  &50.9 & 35.7   &51.4  &63.6    & 67M \\ 
         YOLOv8x  &$1280^{2}$  &NA  &\textbf{65.4} &\textbf{50.8} & \textbf{61.0} &68.8 &68.2M  \\ 
         YOLOv9-e  &$1280^{2}$  &NA  &59.0   &47.0  &52.3   &81.3   &58.1M  \\ 
         YOLOv10-x  &$1280^{2}$  &NA  &52.5   &42.6  & 49.2   &\textbf{85.0}   &29.5M  \\ 
         YOLO11-x  &$1280^{2}$  &NA  &57.8   &44.1  & 60.3   &72.0   &56.9M  \\ 
                          
        \hline
         YOLOv8x  &$2048^{2}$  &NA  &\textbf{69.1}  &\textbf{53.8} &\textbf{66.2 }&78.3  &68.2M  \\ 
         YOLOv10x  &$2048^{2}$  &NA  &53.4   &43.3 &51.0 & \textbf{87.1}  &29.5M  \\ 
         YOLO11-x  &$2048^{2}$  &NA  &68.5 &53.4   &65.2   & 68.4   &56.9M  \\

\hline
\end{tabular}
}
\caption{The tested models on UDD dataset v1 benchmark ($\approx$85\% train/validation, and $\approx$15\% test). Train data: \textbf{6347 images}. Test data: \textbf{1006 images, 10649 instances.}
}
\label{tab:overall_results}
\end{table*}

In our experimental tests, we employ widely recognized performance metrics to evaluate the trained object detection models, including Precision, Recall, mean Average Precision (mAP), and F1-score. The F1-score is a harmonic mean of Precision and Recall, providing a balanced measure of model performance. The mAP is typically reported at an Intersection over Union (IoU) threshold of 0.5 (mAP@0.5) or averaged across multiple thresholds (mAP@[0.5:0.95]). Scores for these metrics range from 0 to 1 (or 0-100\%), with values closer to 1 indicating superior performance.

Our new benchmark dataset, referred as UDD (Usine de Demain), was developed alongside the data acquisition process. Consequently, we evaluate the models across four versions: UDD v1, UDD v2, UDD v3, and UDD v4. The number of images for both training and testing was progressively increased, with an 80-90\% split for training/validation and the remaining 10-20\% reserved for testing. The test images for each version were manually selected to ensure comprehensive validation across all classes and a diverse range of situational scenarios. For all tested models, PyTorch was employed as the testing framework, with default optimizer settings (primarily Stochastic Gradient Descent, SGD) applied to each model. During training, data augmentation techniques, such as vertical and horizontal flipping, along with the mosaic technique, were utilized to improve model performance and robustness. Each model was trained for 500 epochs with an early stopping patience of 100 epochs, and the best-performing version was retained for testing (trained and tested on Nvidia A100 80GB GPU). 

\subsection{Overall performance}
\label{sec:overall_performance}
\begin{figure}[ht!]
    \centering
    \includegraphics[width=0.6\linewidth]{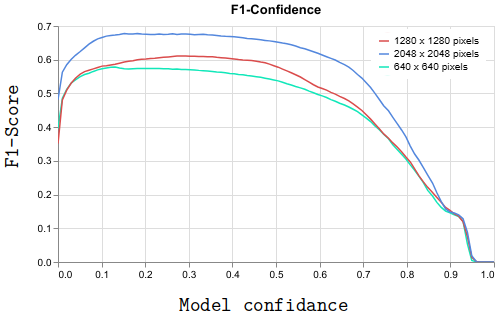}
    \caption{F1-confidence curve on different image resolution using UDD v1 and YOLOv8x. An optimal confidence threshold for deployment is approximately 0.15.}
    \label{fig:F1-score}
\end{figure}

Table \ref{tab:overall_results} presents a comparative analysis of various models across multiple image sizes, along with associated parameters. The models evaluated include several YOLO versions (YOLOv5, YOLOv6, YOLOv7, YOLOv8, YOLOv9, YOLO-NAS, YOLOv9 and YOLO11), Faster R-CNN-R101, and RT-DETR \cite{zhao2024detrs} variants. The models are tested with images of different sizes, ranging from $640^{2}$ to $2048^{2}$, which allows for a broader understanding of performance scaling with input resolution. 

Across the $640^{2}$ resolution, YOLO11-x stands out with the highest mAP@0.5:0.95 (46.7) and mAP@0.5 (62.8), surpassing the performance of YOLOv8, YOLOv9, and earlier models, which all hover around the low 40s in mAP@0.5:0.95. RT-DETR-L1 and RT-DETR-x demonstrate lower performance compared to YOLO models at the same resolution, with mAP@0.5:0.95 values of 43.0 and 44.4, respectively. Their mAP@0.5 scores are also below those of the highest-performing YOLO models. YOLOv8-x consistently exhibits high precision (87.4) across all resolutions, but its recall is lower compared to some other models, such as YOLO11-x. Faster R-CNN-R101, a well-known model, achieves the lowest mAP@0.5:0.95 (44.0) and mAP@0.5 (55.8) at $640^{2}$, further demonstrating that the YOLO series (particularly versions 10 and 11) achieves significantly higher performance in comparison, especially in high-resolution settings.

As image size increases to $1280^{2}$ and $2048^{2}$, performance generally improves as shown in Figure \ref{fig:F1-score}. For instance, YOLOv10-x at $2048^{2}$ achieves an mAP@0.5:0.95 of 53.4 and a notable 69.1 in mAP@0.5, which is the highest across the table for those metrics.  YOLOv11-x, while having fewer parameters (56.9M) compared to YOLOv10-x (68.2M), maintains competitive performance with 68.5 in mAP@0.5 and 53.4 in mAP@0.5:0.95 at $2048^{2}$. The YOLO-NAS-1 model offers a competitive mAP@0.5:0.95 (44.2) while keeping parameter size relatively low (68.2M), showing a balance between accuracy and model complexity. YOLOv10-x at $2048^{2}$ resolution stands out with the highest recall (66.2), indicating the model's effectiveness in detecting a larger number of objects.
Precision generally remains high across the board, with most YOLO models exceeding 85\% at higher image sizes. 

The number of parameters (presented in millions in Table \ref{tab:overall_results}) varies significantly. For instance, YOLOv5x and YOLOv6-l6 have larger parameter counts (140.7M), but their performance does not scale proportionally with this complexity, particularly compared to smaller models like YOLOv11-x with just 56.9M parameters, which delivers superior mAP scores. YOLOv10-x, with 68.2M parameters at $2048^{2}$, appears to strike a strong balance between model size and performance, achieving the highest mAP@0.5 while maintaining reasonable complexity.
  
Performance improves with increasing image resolution, as observed in all models except the YOLO11, which surprisingly has a performance loss with higher resolution. This trend aligns with the expectation that larger input sizes allow models to capture finer details, contributing to better detection accuracy. YOLO models generally outperform both Faster R-CNN and RT-DETR models in terms of mAP@0.5:0.95 and mAP@0.5, while also being more parameter-efficient in several cases. The YOLOv10 and YOLOv11 series, particularly the -x variants, demonstrate the best trade-off between recall, precision, and parameter count, making them the top performers in this table.

\begin{figure}[ht!]
    \centering
    \includegraphics[width=0.8\linewidth]{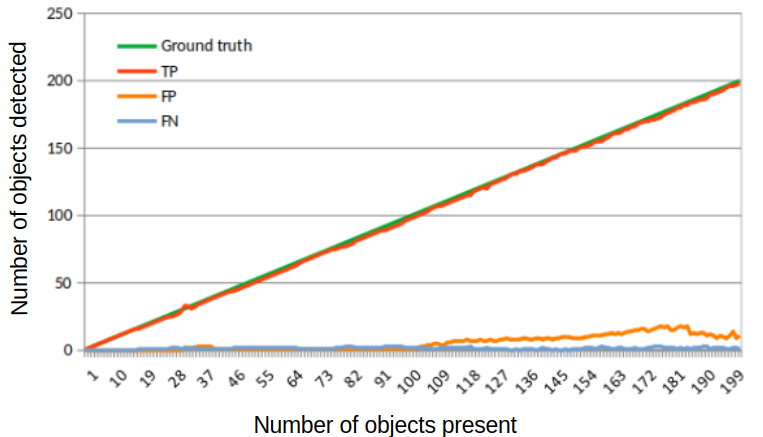}
    \caption{Detection performance exhibited a steady increase as the number of objects rose from 1 to 200. While false negatives (FN) experienced a slight uptick after the 100th instance, false positives (FP) demonstrated a more pronounced increase with the growing number of instances.}
    \label{fig:test_s1}
\end{figure}

Furthermore, we tested the YOLOv8x model on-site evaluation by incrementing the normal class object one by one and recording the true detection TP, the false detection FN, and the false positive detection FP as shown in Fig. \ref{fig:test_s1}. The object detection system demonstrates robust performance when operating in scenes with a limited number of objects. As the object density increases, the system exhibits a corresponding rise in true positives detection, accompanied by a growing incidence of false positives. This behavior is characteristic of object detection systems and reflects the inherent trade-off between true positive and false positive rates. In this particular evaluation, the detection accuracy reached approximately 98\%. The accompanying figure illustrates the system's effectiveness in detecting objects while also highlighting its susceptibility to false positives, particularly in scenarios with high object densities. Field testing revealed an accuracy of approximately ±10 hulls on the simulation table.

The YOLOv8 model introduces several architectural improvements that significantly enhance its performance in small object detection. Specifically, the optimized feature pyramid network (FPN) and path aggregation network (PAN) are designed to preserve fine-grained spatial details across multiple scales, which is essential for accurately detecting small and densely packed objects. Furthermore, YOLOv8 adopts an anchor-free detection mechanism, allowing the model to better localize small objects without relying on predefined anchor boxes that are often ineffective for fine-scale features. In contrast, YOLOv11 prioritizes inference speed by simplifying its neck and backbone structures, reducing computational complexity at the expense of detailed feature extraction. This trade-off makes YOLOv11 less effective in scenarios where precise detection of small objects is required, giving YOLOv8 a clear advantage in such tasks. Overall, since the YOLOv8x emerged as the top performer on our UDD v1 dataset benchmark, we have selected this model for further investigation in the following subsequent sections [\ref{sec:synthetic_and_data_augmentation}, \ref{sec:Histogram_equalization}, \ref{sec:boosting_performance}]. 


\subsection{Synthetic data and data augmentation}
\label{sec:synthetic_and_data_augmentation}
Table \ref{tab:synthetic_results} and Table \ref{tab:synthetic_on_real_imgs} present the performance of the YOLOv8x model trained on synthetic data (V1 and V2) and evaluated on both synthetic and real datasets. While the model exhibited exceptional performance during training and testing on synthetic data, its performance degraded when evaluated on real-world data. This discrepancy can be attributed to the disparity between the colors and textures of synthetic images and those encountered in real-world scenarios. This observation underscores the inherent challenge of generating synthetic data that accurately replicates the dynamic and variable conditions prevalent in industrial environments, such as fluctuations in lighting and projector settings.

Table \ref{tab:data_augmentation} presents the performance of YOLOv8 models trained on the UDD v2 dataset with data augmentation techniques designed to augment the training set through horizontal and vertical flipping. The results unequivocally demonstrate the effectiveness of data augmentation in enhancing model performance. Across all YOLOv8 model sizes, data augmentation consistently leads to improvements in both mAP@0.5 and mAP@0.5-0.95, underscoring its efficacy in improving model performance compared to evaluations on the UDD v1 dataset. However, we investigate the effectiveness of histogram equalization in the following subsection \ref{sec:Histogram_equalization}.

\begin{table}[ht!]
\centering
\resizebox{1\columnwidth}{!}{%
\begin{tabular}{|c|c||c|c|c|c|c|c|}
\hline
    Model & Img size   & mAP@0.5 & mAP0.5-0.95 & Recall& Precision & Parameters\\
\hline                     

         Synthetic v1 &$2048^{2}$ & 94.0 &72.6 &90.9   &93.7   &68.2M  \\              
         Synthetic v2 &$2048^{2}$   & 96.9 &76.7  &93.8    &96.5    &68.2M  \\      
  
\hline
\end{tabular}
}
\caption{The tested YOLOv8x on UDD synthetic dataset. V1 ($\approx$65\% train/validation and $\approx$35\% test): Train data: 24367 images, 4.9M instances. Test data: 12992 images, 1.9M instances. V2: Train data: 12992 images, 1.9M instances. Test data: 24367 images, 4.9M instances.
}
    \label{tab:synthetic_results}
\end{table}

\begin{table}[ht!]
\centering
\resizebox{1\columnwidth}{!}{%
\begin{tabular}{|c|c||c|c|c|c|c|c|}
\hline
    Model & Img size   & mAP@0.5 & mAP0.5-0.95 & Recall& Precision & Parameters\\
\hline                     

         Synthetic v1 &$2048^{2}$ & 39.9 &27.1 &42.8   &51.7   &68.2M  \\              
         Synthetic v2 &$2048^{2}$   & 41.1 &29.5 &44.8    &52.5    &68.2M  \\      
  
\hline
\end{tabular}}
\caption{The tested YOLOv8x trained on UDD synthetic dataset and tested on real dataset UDD v1 (test data: 1006 images, 10649 instances).}
    \label{tab:synthetic_on_real_imgs}
\end{table}

\begin{table}[ht]
\centering
\resizebox{1\columnwidth}{!}{%
\begin{tabular}{|c|c|c||c|c|c|c|c|c|}
\hline
    Model & Img size  &$mAP_{COCO-Val-17}$ & mAP@0.5 & mAP0.5-0.95 & Recall& Precision & Parameters\\
\hline                     
         YOLOv8m  &$640^{2}$  &50.2  &62.1 & 44.2 &60.2  & 79.8 & 25.9M\\             
         YOLOv8l  &$640^{2}$  &52.9  &67.8 &49.6 &65.7 &70.4  & 43.7M\\ 
         YOLOv8x  &$640^{2}$  &53.9  &68.0 &48.7 &66.9 & 70.8 &68.2M \\
          \hline
         YOLOv8x  &$1280^{2}$  &NA  &73.8 &53.1 & 69.6 & 86.4 &68.2M  \\ 
         YOLOv8x  &$2048^{2}$  &NA  &\textbf{78.3} &\textbf{58.1} &76.2  &80.4   &68.2M  \\ 
                           
                           
         \hline
                
                           
\hline
\end{tabular}}
\caption{The tested YOLOv8 on dataset v2 using data augmentation (50\% probability of horizontal flip
and 50\% probability of vertical flip). \textbf{Train data: 14408 images}. Test data: \textbf{1165 images, 11142 instances}.}
    \label{tab:data_augmentation}
\end{table}

\subsection{Histogram equalization}
\label{sec:Histogram_equalization}
Histogram equalization \cite{pizer1987adaptive} is a well-known technique used to enhance the contrast of images by effectively redistributing the intensity values across the entire range of possible values. Therefore, histogram equalization is a valuable pre-processing technique that can significantly improve object detection performance by enhancing image contrast, thereby aiding detection algorithms in better distinguishing objects from their backgrounds. The following steps outline the mathematical formulation histogram equalization:
\begin{enumerate}
\item Calculate the Histogram:
Let \( h(i) \) be the histogram of the image, where \( i \) is the intensity level (from 0 to 255 for an 8-bit image):
\[
h(i) = \text{number of pixels with intensity } i
\]

\item Compute the Cumulative Distribution Function (CDF):
The cumulative distribution function \( C(i) \) is calculated as follows:
\[
C(i) = \sum_{j=0}^{i} h(j)
\]

\item The CDF is then normalized to the range [0, 1] by dividing by the total number of pixels \( N \):
\[
P(i) = \frac{C(i)}{N}
\]

\item Map Intensity Levels:
The new intensity level \( s \) for each original intensity level \( r \) is determined by:
\[
s = \text{round}( (L-1) \cdot P(r) )
\]
where \( L \) is the number of intensity levels (256 for an 8-bit image).
\item Construct the Equalized Image:
Finally, the new image is constructed by replacing each pixel's intensity \( r \) with its corresponding new intensity \( s \).
This process effectively redistributes the pixel intensities, enhancing the overall contrast of the image.
\end{enumerate}

\begin{figure}[ht]
    \centering
    \includegraphics[width=0.9\linewidth]{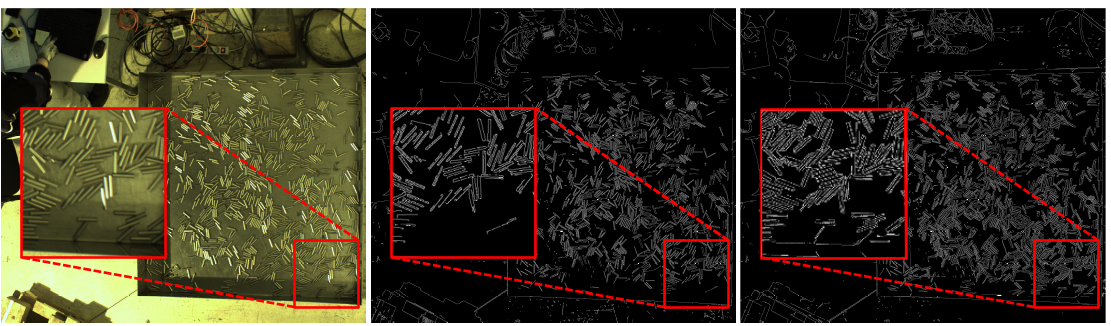}
    \caption{Canny edges before and after Histogram Equalization (HE). Left: original image, middle: canny filter before HE, and right image: canny filter after HE.}
    \label{fig:histogram_equ}
\end{figure}

Figure \ref{fig:histogram_equ} provides a comparative analysis of Canny edge detection \cite{canny1986computational} applied to an image before and after histogram equalization. This visual representation underscores the efficacy of histogram equalization as a pre-processing step for enhancing the performance of object detection models. By effectively increasing the image's contrast, histogram equalization facilitates more accurate and robust edge detection, which is a critical component of convolutional layers in models. This is particularly advantageous for images characterized by low contrast or uneven illumination, where histogram equalization can mitigate the challenges posed in industrial environments. The ablation study presented in Table \ref{tab:histogram_equalization} strongly supports the integration as a pre-processing step within object detection systems. Our findings reveal a notable improvement of 7\% in mAP@0.5 at a resolution of $1280^{2}$ pixels and a 3\% increase at a resolution of $2048^{2}$ pixels.

\begin{table}[ht!]
\centering
\resizebox{1\columnwidth}{!}{%
\begin{tabular}{|c|c|c||c|c|c|c|c|c|}
\hline
    Model & Img size  &$mAP_{COCO-Val-17}$ & mAP@0.5 & mAP0.5-0.95 & Recall& Precision & Parameters\\
\hline                     

         Without HE &$1280^{2}$  &NA  &56.1  & 38.5 & 53.0   &81.8    &68.2M  \\        
 
         With HE &$1280^{2}$  &NA  &\textbf{63.3}  &\textbf{44.4} &62.6   &74.3   &68.2M  \\ 
         
         \hline       
         
         Without HE &$2048^{2}$  &NA  &64.8 &46.4 &63.8  &68.2   &68.2M  \\                    
         With HE &$2048^{2}$  &NA  &\textbf{66.6} &\textbf{46.8} &63.4 &71.3 &68.2M \\   
         
\hline
\end{tabular}
}
\caption{The tested YOLOv8 on dataset v3 (Histogram Equalization test). Dataset ($\approx$80\% train/validation, and $\approx$20\% test). \textbf{Train data: 8086 images}. Test data: \textbf{2066 images, 54747 instances. 
}}
    \label{tab:histogram_equalization}
\end{table}

\subsection{Pre-training on Synthetic dataset}
\label{sec:pre_training_synthetic}

We conducted another ablation study to evaluate the effectiveness of pre-training on synthetic data for improving the accuracy of object detection in specific classes. Table \ref{tab:my_best_model_real} presents the performance of the YOLOv8x model trained exclusively on real data, demonstrating its proficiency in detecting objects from the "normal" class while encountering challenges with objects from the "spring" class.

Table \ref{tab:my_best_model_boosted} showcases the performance of the YOLOv8x model pre-trained on synthetic data (specifically the "normal" class) and subsequently fine-tuned on real data. Our findings indicate that this approach leads to marginal improvements in detecting the target class and similar classes, such as "normal," "pinched," and "smashed." However, it is accompanied by a slight decrease in performance for the remaining classes.

\begin{figure}[ht!]
    \centering
    \includegraphics[width=\linewidth]{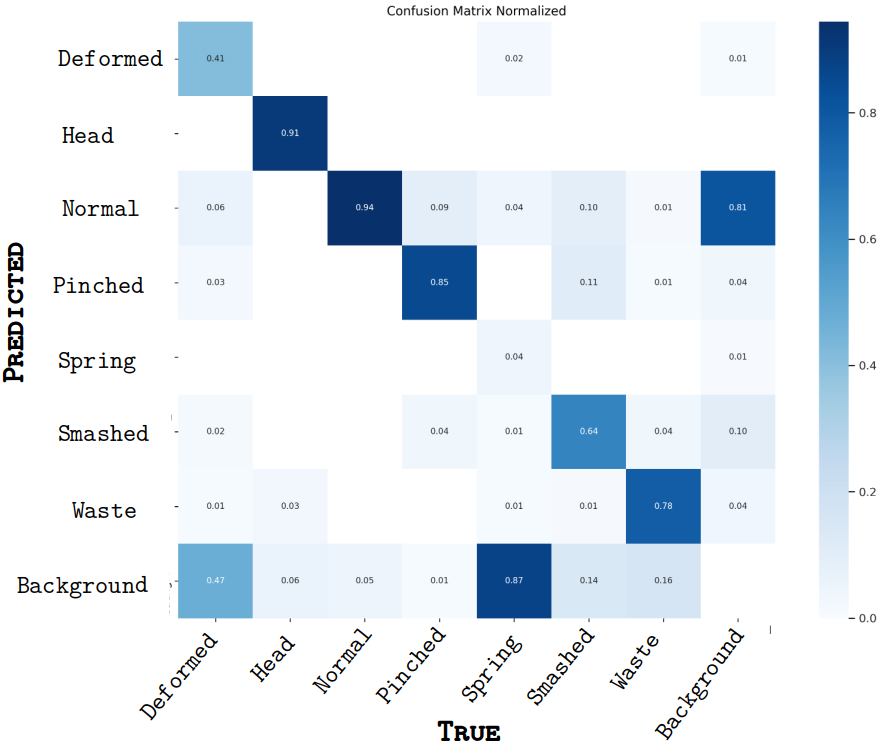}
    \caption{Confusion matrix for model trained only on real data.}
    \label{fig:confusion_real}
\end{figure}

\begin{figure}[ht!]
    \centering
    \includegraphics[width=\linewidth]{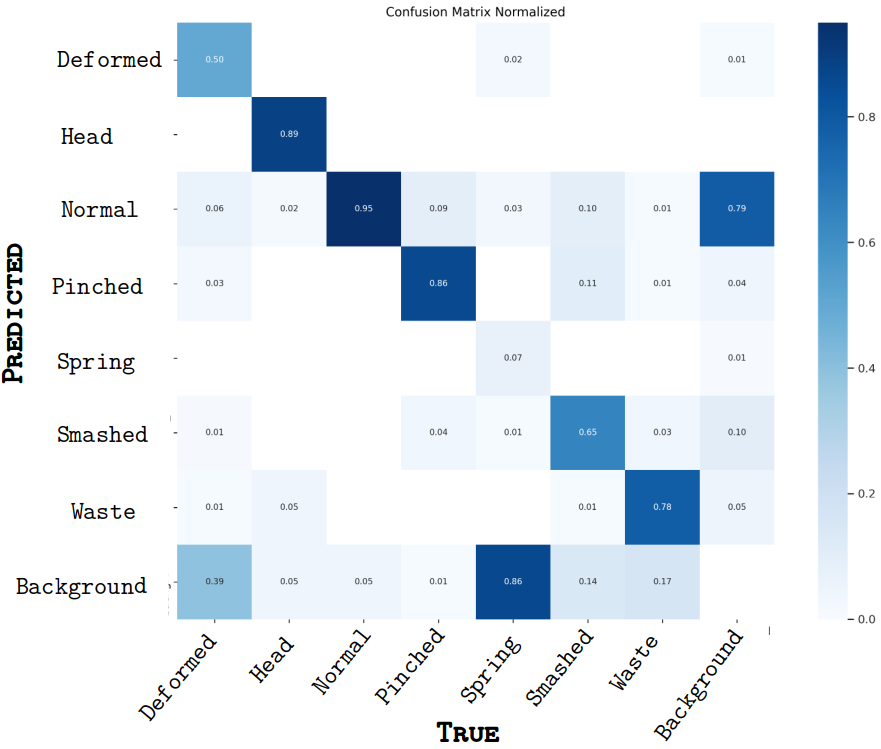}
    \caption{Confusion matrix for model pre-trained on synthetic data and fine-tuned on real data.}
    \label{fig:confusion_boosted}
\end{figure}

The normalized confusion matrices depicted in Figure \ref{fig:confusion_real} and Figure \ref{fig:confusion_boosted} corroborate our earlier conclusion that pre-training on synthetic data can indeed enhance the detection accuracy of specific classes, but this often leads to a trade-off in terms of reduced performance for other classes. 
In this particular experiment, we observed a 1\% increase in accuracy for the targeted classes but a corresponding 1\% decrease in other classes, with a more pronounced 2\% decline for the "deformed" class. Therefore, the decision to employ pre-training on synthetic data should be carefully considered based on the specific requirements and priorities of the application.

\label{sec:boosting_performance}

\begin{table}[ht!]
\centering
\resizebox{1\columnwidth}{!}{%
\begin{tabular}{|c|c|c||c|c|c|c|c|}
\hline
    Model  &Class &Instances  &Precision & Recall &mAP@0.5 & mAP0.5-0.95\\
\hline                             
         YOLOv8x  &normal  &50361 &91.1& 95.0& 96.4& 75.3 \\              
         YOLOv8x  &deformed & 239 &62.2& 48.2& 56.7& 33.6  \\       
         YOLOv8x  &pinched & 1487 &66.7& 88.5& 80.1& 53.9 \\               
         YOLOv8x  &smashed & 2184 &69.5& 67.9& 67.6& 36.8 \\               
         YOLOv8x  &spring  &96 &11.6 & 5.21 & 5.22  &3.27 \\               
         YOLOv8x  &head  &109  &99.0 &91.3 &96.0  &91.7 \\              
         YOLOv8x  &waste &1341  &80.1 &79.3 &82.2  &57.0 \\     
         \hline
         YOLOv8x  &all  &55817 &\textbf{68.6}  &\textbf{67.9}  &\textbf{69.2}  &\textbf{50.2}  \\                      
         
\hline
\end{tabular}}
\caption{Trained only on real data UDD v4. Dataset ($\approx$80\% train/validation, and $\approx$20\% test): 10,438 images (121,663 instances). Train data: 8378 images (65846 instances). Test data: 2060 images (55817 instances).}
    \label{tab:my_best_model_real}
\end{table}

\begin{table}[ht!]
\centering
\resizebox{1\columnwidth}{!}{%
\begin{tabular}{|c|c|c||c|c|c|c|c|}
\hline
    Model  &Class &Instances  & Precision& Recall &mAP@0.5 & mAP0.5-0.95\\
\hline                             
         YOLOv8x  &normal  &50361 &92.0  &95.1 &96.5 &75.7\\              
         YOLOv8x  &deformed  &239 &65.4 &49.1 &54.8 &32.0\\         
         YOLOv8x  &pinched  &1487     &68.7   &88.1 &80.4 &53.9\\            
         YOLOv8x  &smashed   &2184   &70.5  &66.5 &68.9  &38.0 \\               
         YOLOv8x  &spring  &96      &18.5     &8.33     &4.64   &2.41\\             
         YOLOv8x  &head  &109        &99.0      &89.7      &95.6     &91.6\\             
         YOLOv8x  &waste &1341      &79.7      &78.3      &82.2    &57.5\\        
         \hline
         YOLOv8x  &all  &55817   &\textbf{70.7}      &\textbf{67.9}       &\textbf{69.0 }   &\textbf{50.1 } \\                  
         
\hline
\end{tabular}}
\caption{Pre-trained on Synthetic dataset, then fine-tuned on real data UDD v4. Dataset ($\approx$80\% train/validation, and $\approx$20\% test): 10,438 images (121,663 instances). Train data: 8378 images (65846 instances). Test data: 2060 images (55817 instances).}
    \label{tab:my_best_model_boosted}
\end{table}

\section{Length measurement of objects}
\label{sec:length_measurement}
One method for estimating object lengths in images involves employing a segmentation model to extract object masks. These masks can then be used to measure pixel-based lengths, which can subsequently be converted to physical units such millimeters or centimeters. However, this approach is computationally intensive and may exhibit sub-optimal performance when dealing with small, densely overlapping objects. As a result, we opt to utilize the diagonal of bounding boxes as a proxy for hull length. This method offers a balance of speed, computational efficiency.

First, the object is enclosed within a bounding box, defined by its top-left corner at \((x_{\text{min}}, y_{\text{min}})\) and bottom-right corner at \((x_{\text{max}}, y_{\text{max}})\). The diagonal of this bounding box can be computed using the Euclidean distance formula:
\[
d = \sqrt{(x_{\text{max}} - x_{\text{min}})^2 + (y_{\text{max}} - y_{\text{min}})^2}
\]
This diagonal length \(d\) gives a quick approximation of the length of the objects including metallic hull. The diagonal provides an estimate of the object's length for many shapes, as the length of most objects is roughly proportional to the diagonal of their bounding box. However, while this approach is fast, it is not always the most accurate. For irregularly shaped objects, the bounding box diagonal may not closely match the actual length of the object. In such cases, a more precise method involves object segmentation.


\begin{table}[ht!]
\centering
\resizebox{1\columnwidth}{!}{%
\begin{tabular}{|c|c|c|c|c|}
\hline
    Length in mm & Number of reps & TP &Detection accuracy \%  & Measurement Mean Error (+/- mm) \\
\hline                     

         65  & 30 &7 &23,33 &1   \\              
         70 & 30 &9 &30,00&3   \\              
         75 & 30 &15	&50,00 &9   \\              
         90 & 30 &13	 &43,33&4   \\              
         100 & 30 &18 &60,00 &11   \\              
         110 & 30 &23 &76,67&4   \\              
         120 & 30 &28 &93,33 &5   \\              
         130 & 30 &24 &80,00&7   \\              
         140 & 30 &26 &86,67 &7   \\              
         150 & 30 &21 &70,00 &10   \\              
         160 & 30 &24 &80,00 &10   \\              
         170 & 30 &27 &90,00 &11   \\              
         180 & 30 &25 &83,33 &11   \\              
         190 & 30 &27 &90,00 &16   \\              
         200 & 30 &22 &73,33 &20  \\              
         300 & 30 &27 &90,00 &98   \\              
         400 & 30 &27 &90,00 &210   \\              
         600 & 30 &28 &93,33 &377   \\              
\hline
\end{tabular}}
\caption{Detection and measurement accuracy on a stack of 2k objects (metallic hulls), Conducting 30 random repetitions for each length category.}
\label{tab:mesure_error}
\end{table}

\begin{figure}[ht!]
    \centering
    \includegraphics[width=0.9\linewidth]{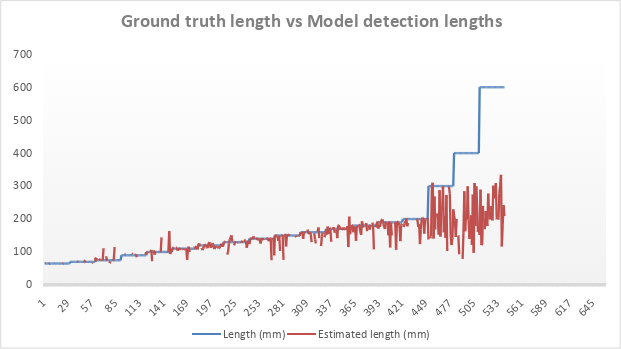}
    \caption{Identify and quantify long metallic hulls within a collection of 2K non-long metallic hulls (classified as "normal"). Conducting 30 random repetitions for each length category.}
    \label{fig:s4}
\end{figure}

On-site evaluation has been performed, we assessed the model's capability to detect 1 to 4 long metallic hulls (e.i., "normal" or class 0) of varying lengths, ranging from 75 mm to 1100 mm. These hulls were introduced into a stack of 2000 hulls through a randomized process, one at a time. The overall detection rate for long hulls reached 72\%. Table \ref{tab:mesure_error} provides a detailed breakdown of the detection performance for each hull length, including the associated measurement error in millimeters (following pixel calibration).

Figure \ref{fig:s4} highlights the model's limitations in accurately detecting long hulls or occasionally misclassifying them as multiple hulls. This behavior can be attributed to the imbalanced distribution of lengths within the training dataset. As we encounter hulls exceeding 300 mm, the model exhibits a tendency to detect them as two or three distinct hulls.

\subsection{Statistical anomaly detection}
\label{sec:statitical_anomaly}
To identify anomalies within the recycling process, specifically large objects and long metallic hulls, we employ the statistical metric known as the Interquartile Range (IQR) \cite{vinutha2018detection}. This approach effectively highlights outliers by focusing on the central 50\% of the distribution, thereby demonstrating resilience to extreme values. A key advantage of using IQR is its independence from camera calibration or reference points, making it robust even when faced with changes in camera zoom or image resolution. Furthermore, this approach eliminates the necessity for explicit length measurements to identify long hulls or larger objects. The following outlines the procedure for identifying outliers using IQR:

First, the first quartile (Q1) is the 25th percentile, meaning 25\% of the data points are below this value, and the third quartile (Q3) is the 75th percentile, meaning 75\% of the data points are below this value.

The Interquartile Range (IQR) is calculated as:
\[
\text{IQR} = Q3 - Q1
\]
This gives the spread of the middle 50\% of the data. Next, the outlier boundaries are defined as follows:
\[
\text{Lower bound} = Q1 - 1.5 \times \text{IQR}
\]
\[
\text{Upper bound} = Q3 + 1.5 \times \text{IQR}
\]

Data points are flagged as outliers if they fall outside these bounds:
\[
\text{Data point} < Q1 - 1.5 \times \text{IQR} 
\]
or
\[
\text{Data point} > Q3 + 1.5 \times \text{IQR}
\]

\begin{figure}[ht!]
    \centering
    \includegraphics[width=0.8\linewidth]{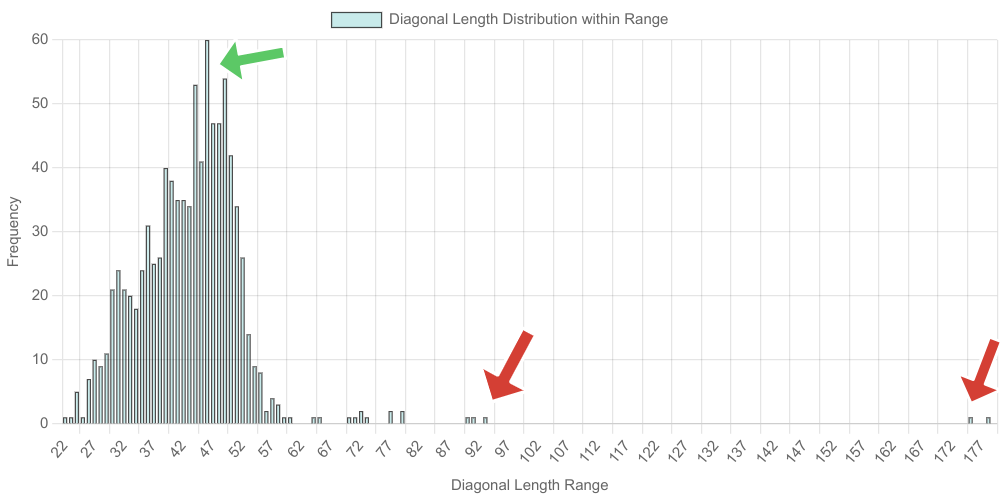}
    \caption{Outliers example in the bounding box diagonal distribution. Green arrows show the median, whereas red arrows indicate outliers.}
    \label{fig:outliers}
\end{figure}

\begin{figure}[ht!]
    \centering
    \includegraphics[width=0.8\linewidth]{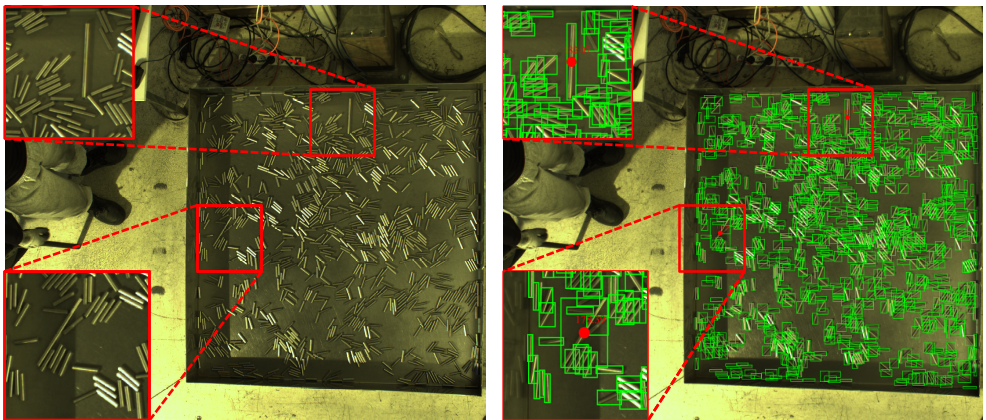}
    \caption{Anomaly detection with IQR over bounding box diagonals. The left image shows the input, while the right image shows the model's output.}
    \label{fig:anomaly_example}
\end{figure}

This method is effective for outlier detection because it is robust to extreme values, as it relies on quartiles rather than the mean or standard deviation. However, given that our application focuses on larger and elongated objects, we exclusively employ upper-bound outliers as indicators of anomalies. Figure \ref{fig:outliers} illustrates the distribution of bounding box diagonal lengths, with outliers identified using the IQR method. Furthermore, Figure \ref{fig:anomaly_example} shows an example of finding long anomaly hulls without the need of length measurement/estimation.

\section{CONCLUSION}
\label{sec:conclusion}
This research presents an in-depth analysis of a novel benchmark database, designated as UDD, specifically addresses the challenges posed by small, dense, and overlapping objects in industrial environment. The experimental results obtained from both on-site and laboratory environments demonstrate that the YOLO family of models, particularly YOLOv8-x and YOLOv11-x, consistently deliver cutting-edge performance, characterized by exceptional mAP scores and precision. The trade-off between model size (parameters) and performance is clearly optimized in these models, making them more suitable for real-world applications where both accuracy and efficiency are critical. This study intends to highlight the pressing need of breakthrough models tailored for such use cases (e.g., industrial environments, manufacturing, or recycling operations).

We conclude that data augmentation techniques expand the diversity of the training data, promoting improved generalization and robustness. However, pre-trained on synthetic data comes with cost, especially in unbalance class distribution. By providing a diverse and controlled dataset, synthetic data can help models learn more generalize features and enhance their ability to detect objects in real-world scenarios. In addition, estimating object length using the bounding box diagonal is a quick and reliable method that works well in many cases. However, for more complex objects where higher precision is needed, a segmentation-based approach is preferred, even though it is more computationally demanding.

\section*{Code - Data Availability}
The repository containing the proposed dataset, methods, and evaluation codes used this study can be found at: \url{https://github.com/o-messai/SDOOD}.

\section*{Disclosures}
The authors declare that there are no financial interests, commercial affiliations, or other potential conflicts of interest that could have influenced the objectivity of this research or the writing of this paper.

\section*{Acknowledgment}
\label{sec:acknowledgment}
This research was conducted as part of the UDD project (Usines De Demain) of Orano group project in collaboration with Bpifrance, this association has pivotal role to advance object detection technologies in complex industrial environments. Therefore, the authors express their deepest gratitude to Orano group and Bpifrance for their crucial support and funding. Special thanks are also due to Siléane group for providing the essential data that formed the basis of this study. Furthermore, we would like to express our sincere appreciation to the Roboflow team \cite{dwyer2022roboflow} for their platform, which greatly facilitated dataset management, annotation.

\bibliography{report}   
\bibliographystyle{spiejour}   





\end{document}